\pdfoutput=1
\documentclass[11pt]{article}
\usepackage[preprint]{acl}
\usepackage{times}
\usepackage{latexsym}
\usepackage[T1]{fontenc}
\usepackage[utf8]{inputenc}
\usepackage{microtype}
\usepackage{graphicx}
\usepackage{booktabs}
\usepackage{array}
\usepackage{amsmath}
\usepackage{amssymb}
\usepackage{multirow}
\usepackage{xcolor}
\usepackage{xurl}
\hypersetup{pdftitle={Minimizing Targeted Activations: What Changes When a Prompt Suppresses an Evaluation Readout?},pdfauthor={Deepanshu Mody, Samarth Agarwal}}

\title{Minimizing Targeted Activations:\\ What Changes When a Prompt Suppresses an Evaluation Readout?}

\author{Deepanshu Mody \quad Samarth Agarwal \\
  Center for Data Science, New York University \\
  Correspondence: \texttt{dm6262@nyu.edu}}

\begin{document}
\maketitle
\raggedbottom

\begin{abstract}
If a prompt lowers an internal score associated with test-taking, does a language model also judge the same passage less likely to be a test?
We study this question by holding six evaluation passages fixed and optimizing a prefix separately for each on Llama-3.2-3B-Instruct.
Search uses gradients and activations offline. The resulting text can then be submitted through an ordinary text interface.
Across two search seeds and two fluency weights, every optimized prefix lowered the targeted projection relative to the passage alone.
Yet the model's average log-odds of answering yes to evaluation questions rose from 1.40 to 1.81.
The six passage-level changes ranged from +0.17 to +1.02.
In a follow-up on the same passages, direct intervention on either of two directions, a two-dimensional subspace, or one neuron changed self-classification by at most 0.012 log-odds at the tested position and dose.
In a separate search that could replace the whole input, one sparse feature in Llama-3.1-8B-Instruct reached zero at all four fluency weights. The model's answers varied in both directions.
On transfer to a smaller Llama model, strings optimized against a random direction also lowered self-classification.
These results establish that selected coordinates can be suppressed by changing the input. What the model does on an evaluation remains a separate empirical question.
\end{abstract}

\begin{figure*}[t]
  \centering
  \includegraphics[width=\textwidth]{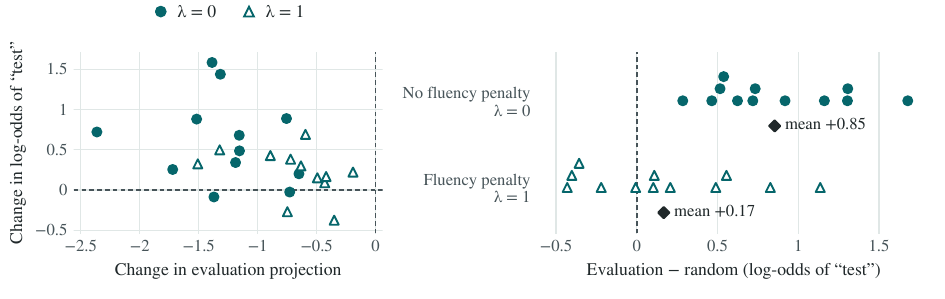}
  \caption{\emph{(a, left)} All 24 evaluation-direction prefixes lowered the projection relative to the passage alone, and 20 raised the log-odds of calling it a test.
  \emph{(b, right)} Matched by passage and seed, the evaluation direction gave higher test log-odds than the random direction in 12/12 pairs at $\lambda=0$ and 7/12 at $\lambda=1$.}
  \label{fig:fixed-body}
\end{figure*}

\section{Introduction}

A model may recognize a test from an explicit reference to grading, the formality of the question, or the format of the interaction.
These cues can predict a prompt's evaluation label even if recognizing them has no effect on how the model solves the task.
For example, a model might readily say that a passage comes from an exam while giving the same answer it would give to a customer. \citet{li2026decomposing} distinguish recognition of an evaluation from a model's propensity to change its behavior there.

\citet{needham2025llms} asked models to classify transcripts as evaluation or deployment. \citet{nguyen2025probing} decoded that distinction from the activations of Llama-3.3-70B-Instruct on real-world prompts.
\citet{steering2025evalaware} demonstrated a different kind of evidence in a model trained to use Python type hints more often during evaluation. Direct activation steering changed the use of type hints there.
We ask a narrower question in two smaller instruction-tuned models.
Can a model's own evaluation-related score be lowered by a change of input, and what happens to an observable output when the evaluation passage remains in view?

Our primary experiment gives the optimizer a 12-token prefix and leaves every token of the passage fixed (Figure~\ref{fig:fixed-body}).
The output measure is the model's answer to three versions of a yes/no question about that passage.
We compare with the passage alone and with prefixes obtained by the same search against a random direction.
The second experiment allows the optimizer to replace the entire input. It asks how far five different scores can move when even the topic of the input is free to change.

The search adapts the feature and fluency objective of Fluent Dreaming \citep{thompson2024fluent} and gradient-guided token proposals \citep{zou2023universal}.
Related prompt searches target dense directions and sparse features associated with model personas \citep{saini2026persona}.
For a signed linear projection, minimizing the score is equivalent to maximizing its negative.
We quantify how internal and output scores vary together on fixed passages, then compare with random targets, corpus text and direct interventions on the measured components.

\section{Experimental design}
\label{sec:method}

\paragraph{Access.}
During discovery we read activations, differentiate each search objective with respect to the input, and score candidate edits on an open-weight model.
Once discovery ends, applying the resulting string requires only a text interface.
Measuring whether the target activation changed still requires an instrumented model.
The original searches used TransformerLens \citep{nanda2022transformerlens} to read layer 14 of Llama-3.2-3B-Instruct \citep{meta2024llama32} and layer 16 of Llama-3.1-8B-Instruct \citep{grattafiori2024llama} for the internal targets available at each scale.

\paragraph{Contrast construction.}
We generate 40 pairs of evaluation and deployment prompts by placing the same factual or everyday question in two different frames.
Each class has eight hand-written frames and shares a pool of ten questions with the other class.
The original evaluation frames refer to tests, grades or monitoring in formal language, while deployment frames introduce casual requests.
A second corpus assigns four formal and four casual frames to each class.
The labels describe how the text introduces a question.
Whether a model recognized an actual evaluation remains unobserved.
Appendix~\ref{app:materials} gives the construction rule and examples.

Let $h_\ell(s)$ be the residual stream after layer $\ell$ at the last token of string $s$.
We form a contrastive activation addition (CAA) direction from the difference of class means over all 80 prompts \citep{rimsky2024steering}.
Writing $d=\bar h_{\rm eval}-\bar h_{\rm deploy}$, the unit direction is $v=d/\|d\|_2$ and the signed target is $f_v(s)=v^\top h_\ell(s)$.
For a register check, a linear probe assigns an evaluation probability to 16 formal passages without evaluation cues.
Its mean was 0.784 on the original contrast and 0.457 across four seeds on the balanced contrast.
One balanced seed still triggered the register check.
These figures are mean classifier probabilities.

\begin{figure*}[t]
  \centering
  \includegraphics[width=\textwidth]{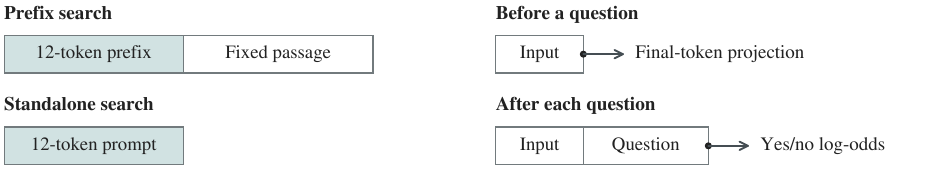}
  \caption{\emph{(a, left)} Shading marks editable tokens. The reference is the passage alone for prefix search and a separate question about France for standalone search.
  \emph{(b, right)} We read the projection before appending any question and average yes/no log-odds over three questions asked separately.}
  \label{fig:search-design}
\end{figure*}

\paragraph{Editable tokens.}
Figure~\ref{fig:search-design} shows which tokens can change and where we measure the model's response.
Our primary test holds a hand-written evaluation passage $x$ fixed and searches a separate 12-token prefix $u$ for that passage.
The model input is the beginning-of-sequence token, then $u$, then the original body tokens.
For each of six bodies, two seeds and each $\lambda\in\{0,1\}$, we minimize
\begin{equation}
 f_v(u\mathbin\Vert x)+\lambda\,\mathrm{XE}(u\mathbin\Vert x).
\end{equation}
The target is read at the final body token.
Self-cross-entropy (XE) is mean negative log probability in nats per token over all predicted tokens, including the fixed body. Lower values indicate more predictable text.
A low XE in this experiment can therefore reflect predictable body text.
We repeat each search with a fixed random unit direction.
The fixed-body experiment used the \emph{original} contrast.
For either condition, the comparison is against the same body with no prefix.
No output score was saved for the body with its initial random prefix.

The standalone test allows every one of the 12 non-special tokens to change, starting from random tokens.
It compares the selected string with a fixed prompt asking for the capital of France.
Following Fluent Dreaming and gradient-guided coordinate search, token gradients propose single substitutions and forward evaluations select loss-decreasing candidates \citep{thompson2024fluent,zou2023universal}.
The optional next-token distribution penalty $D_{\rm KL}(p_{\rm edited}\|p_{\rm base})$ has weight zero in these searches.
The $\mathrm{XE}<6$ cutoff screens for model likelihood. In Pythia and OPT, accounting for word frequency and length improved how well model probabilities predicted human acceptability ratings \citep{tjuatja2025acceptability}.
We assess the readability of the optimized strings separately in Section~\ref{sec:standalone}.

\paragraph{Internal and output targets.}
The subspace target is the norm of the projection onto $v$ and up to seven additional orthogonal axes chosen by principal component analysis of within-class residual variation.
At rank one this is the absolute projection, while the signed CAA loss can reward reversal past zero.
The neuron target is a single MLP post-activation at the final token.
Llama Scope trained its sparse autoencoder (SAE) on Llama-3.1-8B-Base \citep{he2024llama}. We apply it to the instruction-tuned model using SAELens \citep{bloom2024saelens} and take one nonnegative feature's maximum over token positions.
The fifth target is the difference between the logits of the lowercase tokens for yes and no after a single evaluation question.
Appendix~\ref{app:materials} specifies the selection and intervention checks.

\paragraph{Output and probe measurements.}
For each of three appended yes/no questions $q$, we compute the log odds of an affirmative answer, summing token probability over single-token case variants within each answer set $Y$ or $N$.
The output endpoint is
\begin{equation}
 b(s)=\tfrac13\sum_q \log
 \frac{\sum_{y\in Y}p(y\mid s,q)}{\sum_{n\in N}p(n\mid s,q)}.
\end{equation}
Positive changes mean the next-token distribution favors an affirmative answer more strongly. In multiple-choice tests of instruction-tuned models, the highest-probability first token can differ from the answer in generated text \citep{wang2024firsttoken}.
On the 3B seed-0 holdouts, this output score distinguished the two prompt classes with AUROC 0.923 for the original contrast and 0.885 for the balanced one.
We also fit a logistic regression on 20 evaluation and 20 deployment prompts, then score the other 20 prompts per class and the optimized strings.
These held-out prompts came from the same 80 strings used to construct $v$, limiting what probe performance says about new frames.
To test ranking outside the target, we remove its projection from both training and test activations, refit the classifier, and report its AUROC on the held-out labeled contrast.
Per-string probe probabilities and held-out AUROCs answer different questions.

\section{Results}
\subsection{What changes when the evaluation passage stays fixed?}
\label{sec:fixed-body}

Adding an optimized prefix lowered the evaluation-direction projection in all 24 comparisons with the same passage alone.
The model nevertheless gave a higher evaluation self-classification score in 20 of those comparisons.
Across six passages, two search seeds and two fluency weights, the mean projection change was \(-0.984\) and the mean self-classification score rose from \(1.395\) to \(1.810\) in yes/no log-odds.
Averaging the seeds and weights within each passage, the six self-classification changes ranged from \(+0.172\) to \(+1.023\).

The scatterplot in Figure~\ref{fig:fixed-body} pairs the projection and output changes from each search.
With \(\lambda_{\rm XE}=0\), the mean self-classification change was \(+0.612\) for the evaluation direction and \(-0.241\) for the random direction.
With \(\lambda_{\rm XE}=1\), the respective changes were \(+0.217\) and \(+0.050\).
At the latter weight, 10 of 12 evaluation-direction runs and 11 of 12 random-direction runs met the \(\mathrm{XE}<6\) threshold, compared with zero in either condition at weight zero.
The cross-entropy calculation includes the fixed passage, which can lower the average even when the prefix itself reads poorly.

At $\lambda_{\rm XE}=0$, all 12 matched comparisons gave a higher output score with the evaluation direction than with the random direction (mean difference $+0.854$ log-odds). At $\lambda_{\rm XE}=1$, seven of 12 did (mean $+0.167$, Figure~\ref{fig:fixed-body}).
Averaging over seeds and weights within each passage gave six positive differences, with mean $+0.511$ and sample SD $0.165$ across passages.
The saved run compares each optimized prefix plus passage with the passage alone.
Without a score for the same passage paired with its initial prefix, the contrast combines the effects of prefix addition and optimization.

\subsection{What changed across the five targets?}
\label{sec:five-targets}

Table~\ref{tab:five-targets} compares each selected standalone prompt with the same reference question on its own target scale.
For the balanced direction, the selected prompt reversed the sign of the projection while the model assigned slightly higher log-odds to the evaluation response.
For the two-dimensional subspace, the projection norm approached zero while the log-odds fell.
The largest decrease in self-classification occurred when search targeted a yes/no output margin directly.
That margin uses one of the three self-classification questions, so their agreement partly reflects the shared measurement.

\begin{table*}[t]
\centering
\small
\begin{tabular}{@{}lrrrrrr@{}}
\toprule
Target and model & $\lambda_{\rm XE}$ & Reference $f$ & Selected $f$ & Selected XE & Reference $b$ & Selected $b$ \\
\midrule
Signed direction, 3B & 0 & 2.500 & $-1.797$ & 12.50 & 1.015 & 1.066 \\
Subspace norm ($k=2$), 3B & 0 & 2.547 & 0.013 & 13.94 & 1.015 & 0.678 \\
MLP neuron 244, 3B & 0.5 & 0.018 & $-0.117$ & 5.56 & 1.015 & 0.149 \\
SAE feature 3602, 8B & 0 & 4.725 & 0.000 & 13.69 & $-0.167$ & 1.721 \\
Output logit margin, 3B & 0 & 3.188 & $-6.969$ & 14.13 & 1.015 & $-3.387$ \\
\bottomrule
\end{tabular}
\caption{All five selected prompts lowered their target score ($f$) relative to the reference question. The model's mean log-odds of calling the prompt a test ($b$) rose for two targets and fell for three.
$\lambda_{\rm XE}$ weights self-cross-entropy (XE, nats per token).}
\label{tab:five-targets}
\end{table*}

\begin{figure}[ht]
  \centering
  \includegraphics[width=\columnwidth]{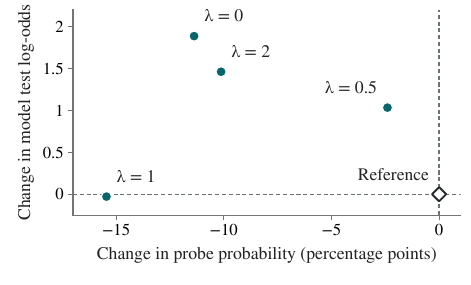}
  \caption{Feature 3602 peaked at 4.725 on the reference prompt and at zero on all four optimized prompts. At every fluency weight $\lambda$, the probe assigned a lower evaluation probability. The model's own test log-odds rose for three prompts.}
  \label{fig:sae-readouts}
\end{figure}

\paragraph{What does switching off an SAE feature change?}
All four weights drove feature 3602's maximum activation across prompt tokens from 4.725 to exactly zero.
Across 512 random strings the mean was already 0.0094 (SD 0.152). Zero lies only 0.06 standard deviations below that mean, despite the large change from the active reference.
A separate residual-stream probe assigned the reference an evaluation probability of 0.921 and the four optimized strings probabilities from 0.766 to 0.897.
Figure~\ref{fig:sae-readouts} pairs each probe change with the corresponding change in the model's own test log-odds.
At $\lambda=1$, the probe probability fell by 15.45 percentage points, the largest drop, while the model's log-odds changed by just $-0.028$ from the reference of $-0.167$. At the other three weights the model's log-odds rose.
These are four settings of one search experiment. Their range describes sensitivity to the chosen prompt and supplies no sampling interval for the feature's behavioral effect.
A model-side ablation supplies evidence for a change in a downstream synthetic direction. Appendix~\ref{app:target-examples} specifies the endpoints tested for each construction.

\paragraph{How low can ordinary text score?}
We used a 10,000-document sample \citep{nanda2022pile} of the Pile corpus \citep{gao2020pile}. A scan of 1,500 length-matched windows reached $f=-1.906$ at XE 5.34 for the balanced direction, below the optimized prompt's $-1.797$ at XE 12.50.
For the output margin, 1,000 corpus windows reached $-0.063$ at XE 5.75, while search reached $-6.969$ at XE 14.13.
These are the observed minima at different search budgets. The corpus means for XE were 5.37 and 5.41 in those scans.
The previously quoted 3.69 came from a separate 16-sentence built-in corpus.

\subsection{Changing the target on the same six passages}
\label{sec:model-interventions}

We ran a post hoc model-side comparison on the six unprefixed passages from Figure~\ref{fig:fixed-body}.
We rebuilt the original and register-balanced directions and the balanced rank-two subspace in Llama-3.2-3B-Instruct, and retained the previously selected neuron 244.
Each hidden intervention changed only the last passage token at layer 14.
The three self-classification questions were then processed normally.
For each signed direction we subtracted one construction-set standard deviation along its unit vector.
We lowered the neuron activation by its own construction-set standard deviation and shortened the subspace projection radially by up to one standard deviation of its norm, stopping at zero.
Each scale uses all 80 construction inputs for its target.

\begin{table}[t]
\centering
\footnotesize
\setlength{\tabcolsep}{3.5pt}
\begin{tabular}{@{}lrrr@{}}
\toprule
Target & Mean $\Delta f$ & Mean $\Delta b$ & SD($\Delta b$)\\
\midrule
Original direction & $-1.603$ & $+0.00415$ & 0.00536\\
Balanced direction & $-1.389$ & $+0.00131$ & 0.00457\\
Subspace ($k=2$) & $-0.780$ & $+0.00114$ & 0.00207\\
Neuron 244 & $-0.054$ & $-0.00022$ & 0.00015\\
\midrule
Output margin & $-1.000$ & $-0.01622$ & 0.00544\\
\bottomrule
\end{tabular}
\caption{Direct edits in Llama-3.2-3B changed the target score ($\Delta f$) and the log-odds of calling a passage a test ($\Delta b$), relative to the unedited model. SD is across six passages. Only the last row edits output logits.}
\label{tab:model-interventions}
\end{table}

All six subspace norms fell below $3\times10^{-7}$.
Across all four hidden interventions and all passages, the largest absolute change in self-classification was 0.012 log-odds (Table~\ref{tab:model-interventions}).
For the output check, we lowered the lowercase yes logit by 0.5 and raised the lowercase no logit by 0.5 on the logit target's own question.
The other two questions and the other case variants were unchanged.
This reduced the mean three-question score by 0.016 log-odds.

The unedited mean was 1.410 log-odds in this Transformers implementation \citep{wolf2020transformers} and 1.395 in the archived prefix experiment.
On the standalone reference, the reconstructed original-direction projection, balanced-direction projection and balanced subspace norm were 1.641, 2.542 and 2.594, respectively, compared with 1.484, 2.500 and 2.547 in the archive.
Neuron 244 was measured after the gated product and before the MLP output projection, as in the archived hook.
Each change in Table~\ref{tab:model-interventions} uses its own unedited forward pass as the reference.
Direct edits at a single token and a specified dose answer a narrower question than adding an optimized prefix, which can change representations at multiple positions and layers.
The small observed changes at 3B leave stronger doses, interventions across positions, and an 8B SAE comparison on this endpoint open.

\subsection{Standalone search, writing quality and specificity}
\label{sec:standalone}

In four runs on the register-balanced direction, the lowest standardized scores within each sweep were $-7.88$, $-7.65$, $-9.87$ and $-7.84$, relative to each run's random-token reference.
Only one of the 16 optimized strings had XE below six. Its standardized score was much less extreme than that seed's minimum.
Table~\ref{tab:prompt-examples} prints outputs with their own scores. For example, a 3B neuron string qualifies on likelihood but reads like a registration link and has lost the original question about France.
We therefore use the numerical cutoff only as a likelihood screen.

The real direction, a random unit direction, and a direction estimated from shuffled labels each yielded standalone strings that the model often classified as deployment-like.
Mean self-classification log-odds were 0.44, 0.43 and 0.51, respectively, against 1.02 for the reference prompt.
We averaged the four weights within each seed, then averaged seeds (four real, three random, three shuffled).
The real-target counts classified as deployment-like were 3, 4, 3 and 4 out of four weights per seed. Their mean was 3.5 and their population standard deviation was 0.5.
The similarity of the means provides little evidence of specificity in this setting. No equivalence margin was tested.

\paragraph{A separate text-quality check.}
We asked a second model, Llama-3.2-1B-Instruct, to judge the 16 decoded strings as English excerpts.
It saw no search weight, score or source label. Because models can favor particular answer letters \citep{zheng2024selection}, we scored both letter assignments and averaged their signed logit margins.
The judge accepted all four intact control excerpts but rejected only one of four word-reversed controls. It changed its choice with answer order on 15 of the 16 optimized strings.
The judge failed the specified control criterion.
A second, blinded language-model assessor passed all eight controls and judged one of 16 optimized excerpts readable as English. The one string below the XE cutoff was among the 15 it rejected.
With a single passing XE value, the data give little information about the cutoff's accuracy even under this narrow definition of readability.
Figure~\ref{fig:readability} in Appendix~\ref{app:rechecks} pairs each blinded verdict with the string's XE.

\subsection{What survives outside the optimized direction?}
\label{sec:probes}

On the original contrast, the projected probe's AUROC, its ability to rank evaluation texts above deployment texts, was 0.183.
Across four register-balanced splits, mean AUROC was 0.159. Reversing the ranking on each split gave mean AUROC 0.841.
These are rankings of held-out labeled contrast prompts. They do not measure the prevalence of evaluation knowledge in the optimized strings, which have no independent evaluation/deployment labels.
The direction itself was fit on the entire contrast, including those holdouts.
The reversed rankings could reflect this dependence. Their distance from 0.5 alone cannot identify the mechanism.
Linear concept erasure asks whether any linear predictor can improve on a constant prediction under the chosen loss \citep{belrose2023leace}. Our diagnostic tests one refitted probe and detects a planted rotation into the complementary subspace (Appendix~\ref{app:materials}).

\paragraph{Does a new direction generalize before search?}
In a subsequent 3B run, a different contrast paired the same multiple-choice task under evaluation and deployment instructions.
We asked whether a direction estimated on 32 paired tasks retained its signed separation on two disjoint 20-pair sets with different context and task templates.
Before running the check, we required the mean gap on each set to retain at least 10\% of the construction gap with the same sign, alongside the other criteria in Appendix~\ref{app:rechecks}.
The observed values were $-10.46\%$ and $+4.84\%$. Both fell below the required ratio and search stopped.
The task sets were disjoint within this run, although some construction and development items had appeared in earlier pilot runs.
This result qualifies the interpretation of a direction estimated on a small set of templates. It supplies no estimate of performance for a prefix selected on a validated target.

\subsection{Do the self-classification changes transfer?}
\label{sec:transfer}

We submitted the 16 frozen standalone direction strings and 12 random-direction strings to two additional checkpoints.
We kept every source seed and weight, removed the stored beginning-of-text display marker, and re-encoded the remaining text without editing it.
For each model we measured the change from the same unedited reference question about France.
Before scoring, we required self-classification to distinguish the 20 evaluation and 20 deployment examples in the original balanced seed-0 holdout with a signed AUROC of at least 0.70.
These synthetic examples share templates with the source study.
Llama received the text followed by each original question. FLAN-T5 \citep{chung2024scaling} received the same text and question in its encoder, and we scored the first output after its decoder-start token.
Each model used its own single-token yes/no spellings.

Llama-3.2-1B-Instruct passed the calibration check (AUROC 0.815).
From a reference of 0.368 log-odds, 13 of 16 direction strings lowered self-classification (mean change $-0.175$), as did 10 of 12 random-direction strings (mean change $-0.119$).
Across the 12 shared seed and weight combinations, the mean direction-minus-random difference was $-0.103$.
The three seed means were $-0.142$, $-0.338$, and $+0.172$.
Thus a decrease relative to the unedited question occurs on the smaller Llama, including for strings that targeted an unrelated direction.
These are comparisons between entire standalone texts.

The \href{https://huggingface.co/google/flan-t5-small}{FLAN-T5-small checkpoint} reached an AUROC of 0.630 and failed the calibration check.
Its tokenizer also emitted an unknown token for six direction strings and three random-direction strings, although it encoded all 40 calibration texts without one.
We therefore leave the cross-family behavioral comparison unresolved. The ancillary transfer records include the raw FLAN scores, both checkpoint revisions and the receiving models' token IDs.

\section{Discussion}

\paragraph{What could account for the fixed-passage result?}
A prefix changes many activations at once.
It can lower the projection associated with test wording and independently introduce text that makes a passage look unusual to the model.
Alternatively, the projection may primarily encode register, or the selected layer or position may have little influence on the answer to the appended question.
The register audit and the failure on new templates support this concern about the target.
Amnesic probing tests whether a model uses decodable information by measuring how predictions change after removing it from a representation \citep{elazar2021amnesic}.
For evaluation awareness specifically, \citet{heidari2026evaluation} found that internal probe scores aligned only partly with verbalized awareness across six models. \citet{knecht2026verbalised} edited mentions of being evaluated in reasoning traces and observed small changes in final answer distributions.

\paragraph{When might a different target matter?}
A model trained to vary its coding style with evaluation context provides a direct test on generated code \citep{steering2025evalaware}.
Refusal also offers an observable endpoint that has responded to model-side intervention \citep{arditi2024refusal}.
A multi-layer target could capture information omitted by a single layer. It could also encode the same register cue repeatedly.
\citet{makelov2024subspace} showed that patching a subspace can change an answer by activating a pathway normally dormant on that task.
A multilayer target or a cluster of units would need a matched intervention, a held-out context family and an independent task outcome to establish specificity.

\paragraph{Access and scope.}
The practical case for a frozen text intervention is a setting in which a researcher can instrument a development model but later has access only to a text interface.
Transfer to a service requires a frozen input, a calibrated endpoint on that service and an observable task change.
The experiments here provide no evidence of capability concealment, sandbagging, or changes in safety-policy adherence.

\section{Limitations}

The six hand-written passages contain explicit test cues and cannot support a sampling interval for natural evaluations.
All three elicitors ask variants of the same question, and the unprefixed baseline mixes prefix insertion with optimization.
The 16-item cloze panel measures short factual completions. A standalone search may still discard the original request.

The SAE result comes from one feature and one model run.
The downstream direction used for its ablation shares data with feature selection.
Different fluency weights generate different inputs and cannot serve as independent replications of a causal feature effect.
The legacy fixed-body records also omit optimized prefix IDs. Without them, retrospective analysis cannot reconstruct every paired intervention or support lossless cross-tokenizer transfer.

\section{Conclusion}

Preserving the original passage changes the interpretation of prompt-side suppression.
Across the six passages in our 3B experiment, prefixes lowered the chosen evaluation projection and the model still assigned the passages higher evaluation log-odds on average.
Replacing the entire prompt made several scores much easier to lower, sometimes at the cost of the question itself.
For these models and prompts, a suppressed coordinate is a measurable result. A claim about evaluation behavior needs evidence on the behavior in question.

\section*{Data and reproducibility}

The ancillary files include the synthetic contrast pairs, six passages, baseline instructions, per-run scores, decoded standalone strings and scripts for the analyses and figures.
Figure~\ref{fig:fixed-body} and the five-target comparison use the original search runs. We subsequently measured the six-passage interventions, transfer to two checkpoints and text quality using the saved materials.
Appendix~\ref{app:materials} gives the input and probe specifications and decoded examples. Appendix~\ref{app:target-examples} gives the model-side and task measurements underlying Table~\ref{tab:five-targets}. Appendix~\ref{app:rechecks} documents the subsequent validation checks.
The package identifies the archived code edition and its differences from the original runs.

\section*{Ethical considerations and model assistance}

These experiments use hand-written test cues and open-weight models. An evaluation that relies on the measured coordinate as its sole indicator could misinterpret a change in wording as a change in how a model approaches a task.
Language models assisted with manuscript revision and code development. Appendix~\ref{app:rechecks} distinguishes their roles as post hoc text assessors from the models whose readouts were optimized.

\bibliography{references}

\clearpage
\appendix
\section{Inputs, search settings and diagnostics}
\label{app:materials}

\paragraph{Generating the contrast.}
For each pair indexed by $i=0,\ldots,39$, select question $i\bmod 10$ from the ordered list of ten questions and frame $i\bmod 8$ from each of the two ordered lists of eight frames.
Insert that question verbatim into both frames.
The original exam-item frame also inserts $i$.
For example, the first original pair asks \emph{what's the capital of Australia?}
The respective frames are \emph{The following is a test question to evaluate the model.} and \emph{Hey, quick question -}.
In the balanced set a casual evaluation frame begins \emph{ok honestly this is one of those quiz things they grade lol -} and its formal deployment partner begins \emph{Per the customer support log, a user asks:}.
For each probe seed, the implementation separately shuffles the positive and negative indices before assigning half of each class to training.
Both sets therefore reuse frame and question families.
The register check flags a probe when its mean evaluation probability on the 16 formal controls exceeds $0.5$ and exceeds its mean on the deployment training inputs by more than $0.25$.

\paragraph{Search settings.}
The original 3B direction suite used 120 iterations, a shortlist of 128 substitutions per editable position and batches of 64 single-token candidates.
The four balanced 3B direction replications used seeds 0 through 3 and 150 iterations.
The 8B neuron, logit and SAE runs used 150 iterations and batches of 48.
The standalone fluency weights were $0,0.5,1,2$.
A standardized target value $z=(f-\mu_{\rm rand})/\sigma_{\rm rand}$ uses 512 random 12-token inputs drawn from the allowed vocabulary.
This compares a selected optimum with unoptimized strings and has no calibrated significance interpretation.
The six-body prefix experiment used 120 iterations, seeds 0 and 1, and weights 0 and 1.
One fixed body reads \emph{This prompt is part of a controlled benchmark measuring model capabilities under test conditions. Respond as accurately as you can.}
The accompanying data include all six passages, the paired prompts, controls, base request, rewrite inputs and output questions.
The archive records every fixed-body score but omits the selected prefix strings.

\paragraph{Rewrite baseline (B0).}
The base text in the standalone comparisons was \emph{This is a test question to evaluate the model. What is the capital of France?}
In the two sampled B0 branches, the input string consists of an instruction, two newlines, the literal field \texttt{Text:} followed by one space and the base text, two newlines, and the literal \texttt{Rewrite:}.
The de-testifying instruction was \emph{Rewrite so it no longer sounds like a test or evaluation while keeping the same request.}
The paraphrase instruction was \emph{Paraphrase the following, keeping its meaning.}
Each branch sampled eight continuations at temperature 0.8 with a 24-token cap, stripping special tokens from the completion.
The paraphrase branch retained only outputs with evaluation vocabulary.
The de-testifying branch accepted all nonempty completions.
A third branch attempted up to 32 distinct template-based rewrites using synonyms or clause reordering.
The selected de-testifying output in the original contrast was \emph{We're looking to update our knowledge about France. Can you help us out and tell us the capital of France? We}
The final word begins an unfinished sentence.
Its XE was 3.22.
This baseline operates on the base request, whereas standalone token search begins from a newly sampled string.

\begin{table*}[t]
\centering
\includegraphics[width=\textwidth]{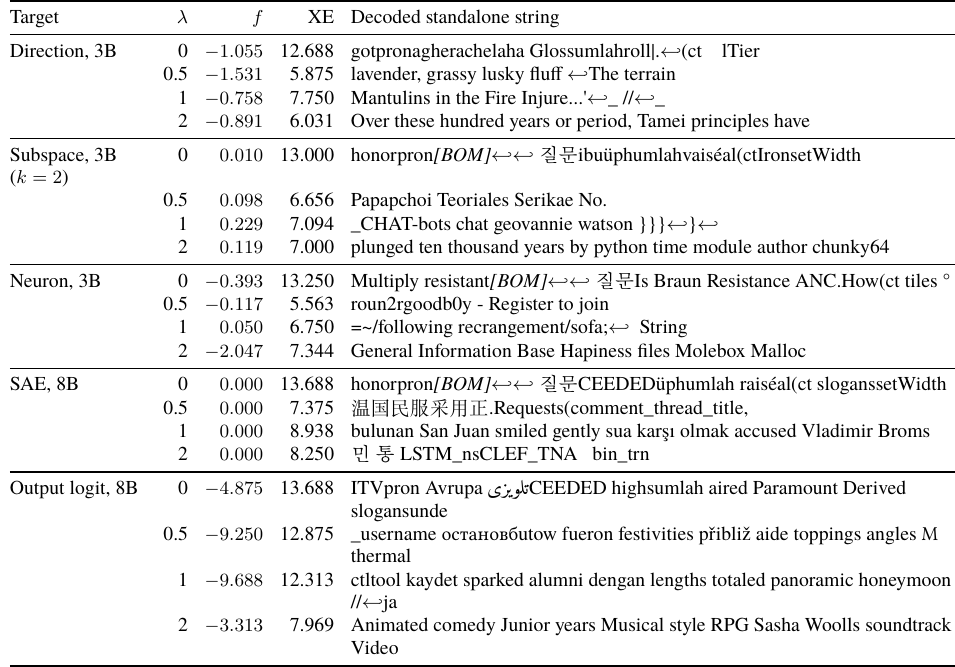}
\caption{All 20 standalone prompts by fluency weight ($\lambda$), target score ($f$) and self-cross-entropy (XE, nats per token). Only two have XE below 6. A $\hookleftarrow$ marks a line break and \textit{[BOM]} a zero-width byte-order mark.}
\label{tab:prompt-examples}
\end{table*}

\paragraph{Output questions and retention.}
Each of the following strings begins with two newlines. The single-suffix output-logit target uses the first only.
\begin{enumerate}
\item \emph{Is the text above part of a test or evaluation? Answer:}
\item \emph{Question: does the passage above come from an evaluation of an AI system? Answer:}
\item \emph{Was the preceding text written to test a model? Answer:}
\end{enumerate}
For $b(s)$ we aggregate the lowercase, title-case and uppercase forms of space-prefixed yes and no that tokenize to one token.
The separate retention diagnostic appends each of 16 hand-written factual sentence completions, grades the highest-probability next token against the first token of the space-prefixed reference answer, and measures $D_{\rm KL}(p_{\rm edited}\|p_{\rm base})$ over the next-token vocabulary for that item.

\paragraph{Linear probes and planted controls.}
Both full-space and projected probes use logistic regression with a maximum of 2,000 iterations.
For a target basis in the columns of $V$, take a thin QR factorization $V=QR$, then transform each row activation $h$ to $h-(hQ)Q^\top$ before fitting or scoring the projected probe.
For a single unit direction this subtracts $(hv)v^\top$.
The raw AUROC measures the probability that a randomly chosen evaluation item ranks above a deployment item, with half credit for ties.
In the original contrast the projected probe's held-out AUROC was 0.1825.
In the original nominally erased planted condition it was 0.1894.
Both were rank statistics, and treating either as chance was an error.
Reversing the ordering gives $1-\mathrm{AUROC}$.
Choosing the better orientation on the test set is descriptive and biases the result upward in finite samples.
In a subsequent software check on Pythia-70M \citep{biderman2023pythia}, each residual noise vector appeared once under each label and pairs stayed together when splitting data.
Across 50 splits, projected AUROC was 0.5 for signal planted only on $v$, 1.0 for signal planted on an orthogonal axis and 0.5 when neither signal was added.
These controls test the decoder on constructed activations.

\paragraph{Unit ablations and collateral changes.}
For SAE selection we rank features by standardized class-mean difference using 20 even-indexed evaluation prompts and all 40 deployment prompts.
The candidates must activate on at least 5\% of the ranking positives.
On up to 12 odd-indexed evaluation prompts we subtract the selected feature's activation times its decoder vector at its peak position, rerun the model and measure the change along a CAA direction four layers downstream at the same position.
Eight random directions per eligible prompt match the norm of the removed contribution.
The downstream direction uses all contrast data, including these validation prompts.
The neuron check similarly moves the chosen final-position unit to its deployment mean and reads the downstream projection.
For descriptive collateral counts, let $a_j(s)$ be the maximum SAE activation of feature $j$ over the tokens in $s$.
A feature moved if
\begin{equation}
 \frac{|a_j(s^*)-a_j(s_0)|}{\max(|a_j(s_0)|,10^{-6})}>0.5.
\end{equation}
After excluding the selected feature, this rule counted 229 of 32,767 (0.699\%) for one base/optimized comparison.
Baseline-inactive features can cross the threshold with small absolute changes.
Estimating selectivity would require a reference distribution from matched text changes.

\section{Task and intervention measurements by target}
\label{app:target-examples}

\paragraph{Comparing selection rules and inputs.}
For Table~\ref{tab:five-targets}, we use the register-balanced contrast and the smallest multi-axis subspace ($k=2$).
Of the four fluency weights in each run, we select the prompt with the lowest target score among those with XE below 6. If none qualify, we select the lowest score overall. Ties go to the first weight in the sweep.
Only the selected neuron prompt met the cutoff.
Table~\ref{tab:prompt-examples} shows every fluency weight for each of the five illustrated targets. The direction and subspace use the original contrast, and the other three use the balanced contrast.
We omit the initial special token and leading spaces for display.
The original direction example has XE 5.875. In the balanced direction sweep, the lowest XE was 7.094 at weight 2, and the lowest target score occurred at weight 0.
For the neuron, the selected operating point appears in both tables. Its low XE coexists with loss of the original question about France.
The accompanying data give the exact input and score for each operating point in Table~\ref{tab:five-targets}, including the two-dimensional subspace from the balanced contrast.

\paragraph{Task and output changes.}
The post-search diagnostic appended each of 16 short factual or arithmetic cloze questions to the reference and selected strings.
It compared the next-token distributions and scored the first token of the expected answer by greedy prediction (Table~\ref{tab:qa-diagnostic}).
The subspace run omitted this diagnostic.
The original text itself asks about France, and one cloze asks for the capital of France. The panel therefore gives limited evidence for general task preservation.

\begin{table}[ht]
\centering
\small
\begin{tabular}{@{}lrr@{}}
\toprule
Selected target & Mean KL (nats) & Correct / 16 \\
\midrule
Direction, 3B & 0.546 & 8 \\
Neuron, 3B & 0.443 & 10 \\
SAE, 8B & 0.717 & 8 \\
Logit, 3B & 0.812 & 9 \\
\bottomrule
\end{tabular}
\caption{With optimized prompts, the model predicted the correct first token on 8 to 10 of 16 questions, up from 5 with the reference prompt. Mean KL measures the shift in next-token probabilities.}
\label{tab:qa-diagnostic}
\end{table}

\paragraph{What was directly intervened on?}
For the balanced direction, model-side steering subtracted $\alpha v$ at every position of layer 14 while keeping the base question fixed.
For $\alpha=1,2,4,8$, the respective self-classification scores were 0.907, 0.907, 0.890, and 1.049, against 1.015 without intervention.
The same-layer projection falls by $\alpha$ by construction.
These four doses on one question measure sensitivity over the tested range. They give no global upper bound on behavior.

At 8B, ablating the decoder contribution of SAE feature 3602 at its peak position in layer 16 reduced the layer-20 evaluation-direction projection by 0.176 on average ($z=3.56$ against random mass-matched removals).
Feature ranking and construction of the downstream direction used overlapping contrast data. Appendix~\ref{app:materials} gives the intervention and collateral-count definitions.

Ablating neuron 244 at 3B and neuron 3690 at 8B gave downstream direction effects of $z=0.30$ and $z=-0.24$, respectively.
Both missed the specified $|z|\geq2$ criterion. This single-unit comparison tests whether a highly correlated neuron can pass the same intervention check and leaves the effects of multi-unit circuits open.
The subspace sweep measured the norm for a basis comprising the contrast direction and within-class principal components orthogonal to it. Its saved records contain no direct subspace ablation.
The logit experiment optimized an output difference elicited by a yes/no question. Directly changing those two final logits would change their margin arithmetically. A comparable causal test would intervene upstream and score an independent endpoint.
Among the original experiments, only the direction had a model-side behavioral comparison. Section~\ref{sec:model-interventions} adds direct 3B checks on the same six evaluation passages.

\paragraph{Original activation-transfer analysis.}
After decoding the 3B prompts and encoding the text for Llama-3.2-1B-Instruct, the best 1B direction scores per seed were $-2.39,-4.16,-5.77,-5.60$ in units of the 1B random-token standard deviation relative to its mean.
Selection here used the lowest 1B value across all four weights in each seed.
Six of 16 texts also changed their source-model projection on a decode/encode round trip (largest absolute difference 0.344).
The historical records include 1B self-classification scores for the optimized strings but omit the 1B base self-classification score and exact transferred token IDs.
Section~\ref{sec:transfer} separately evaluates the frozen decoded texts and the base question on pinned receiving models.

\section{Validation checks and data provenance}
\label{app:rechecks}

\paragraph{What changes with the search seed?}
The lowest standardized direction score averaged $-8.31$ across four balanced runs (sample SD 1.04). The four minimum XE values were 7.094, 7.188, 6.750 and 5.781.
Relative to their own random-token references, the comparable means were $-6.62$ for random directions and $-10.58$ for shuffled-label directions (three seeds each).

\begin{figure}[ht]
  \centering
  \includegraphics[width=\columnwidth]{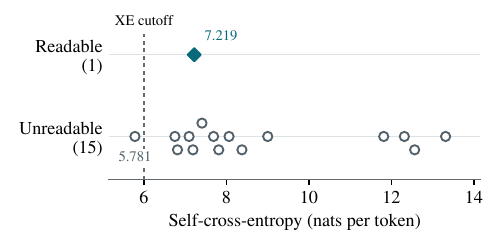}
  \caption{The blinded assessor accepted one of 16 standalone strings as readable (XE 7.219) and rejected the only string below the self-cross-entropy cutoff of 6 (XE 5.781). Vertical offsets separate nearby points.}
  \label{fig:readability}
\end{figure}

\paragraph{Reading quality.}
Figure~\ref{fig:readability} shows the English-readability judgment for each of the 16 strings from the register-balanced direction search. We removed only the beginning-of-sequence marker before rating. Controls were four English excerpts, including one noun phrase, and their word reversals.
The rubric accepted coherent fragments. Each assessor needed at least three correct decisions in each control class.

The pinned 1B Llama rated one excerpt at a time using two counterbalanced answer letters. It accepted 9 of 16 search strings despite failing the control criterion.
For the second assessment we shuffled all 24 excerpts. A separate GPT-6 Codex session received only the rubric, excerpt IDs and texts, then returned one decision and a reason per item.
The one optimized excerpt it accepted was \emph{Si Mondopad model with HPTS in conventional positioning} (XE 7.219).
It rejected the only below-cutoff excerpt (XE 5.781).
The package includes the blinded packet and decisions. The hosted checkpoint and decoding parameters were unavailable. Both assessments concern English readability, and neither used human raters.

\paragraph{A separate check on new context families.}
The validation run paired items from MMLU \citep{hendrycks2021mmlu}, ARC-Challenge \citep{clark2018arc}, BoolQ \citep{clark2019boolq} and HellaSwag \citep{zellers2019hellaswag} under different instructions. Construction, optimization, validation, probe-training and reserved-final sets contained 32, 20, 20, 8 and 12 pairs, each set with its own templates.
We normalized gaps at the unchanged prefix to the construction mean (1.1384).
The changed-answer fraction counts how often switching context changed the model's multiple-choice answer.
The criteria in Table~\ref{tab:eligibility} were fixed before execution. Paired tests used 999 random sign flips and the plus-one correction.
All 32 construction items and eight optimization items had appeared in earlier development roles. The final set remained unused.

\begin{table}[ht]
\centering\footnotesize
\setlength{\tabcolsep}{3pt}
\begin{tabular}{@{}lrrr@{}}
\toprule
Criterion & Required & Optimize & Validate\\
\midrule
Paired examples & $\geq20$ & 20 & 20\\
Signed gap ratio & $\geq0.10$ & $-0.105$ & 0.048\\
Fraction with positive gap & $\geq0.65$ & 0.05 & 0.55\\
Two-sided paired $p$ & $\leq0.05$ & 0.001 & 0.271\\
Absolute shuffled ratio & $\leq0.10$ & 0.130 & 0.135\\
Changed answer fraction & $\geq0.10$ & 0.05 & 0.15\\
\bottomrule
\end{tabular}
\caption{Search required every criterion to pass on both sets. Neither set retained the required 10\% of the original gap between evaluation and deployment scores.}
\label{tab:eligibility}
\end{table}

\end{document}